\documentclass[manuscript]{acmart}

\usepackage{color, colortbl}
\usepackage{amsmath}
\usepackage{array}
\usepackage{keystroke}
\usepackage{makecell}
\usepackage{pifont}
\usepackage{graphicx}
\usepackage{booktabs} 
\usepackage{array}
\usepackage{multirow}
\usepackage[export]{adjustbox}
\usepackage{pifont}
\usepackage{balance}
\usepackage{subfigure} 
\usepackage{tabularx}
\usepackage{menukeys} 
\usepackage{enumitem}
\usepackage{caption}
\usepackage{fontawesome}

\definecolor{materialteal}{HTML}{009688}
\definecolor{materialpurple}{HTML}{9C27B0}
\definecolor{materialindigo}{HTML}{3F51B5}
\definecolor{materialblue}{HTML}{2196F3}

\AtBeginDocument{%
  \providecommand\BibTeX{{%
    \normalfont B\kern-0.5em{\scshape i\kern-0.25em b}\kern-0.8em\TeX}}}

\def \debug{}
\ifx \debug \undefined
    \newcommand{\fx}[1]{{}}
\else
    \newcommand{\fx}[1]{{\textcolor{red}{#1}}}

\newcommand{\xdownarrow}[1]{%
  {\left\downarrow\vbox to #1{}\right.\kern-\nulldelimiterspace}
}

\definecolor{materialcyan}{HTML}{00BCD4}
\definecolor{materialteal}{HTML}{009688}
\definecolor{materialgreen}{HTML}{4CAF50}
\definecolor{materiallime}{HTML}{CDDC39}
\definecolor{materialamber}{HTML}{FFC107}
\definecolor{materialbrown}{HTML}{795548}
\definecolor{materialred}{HTML}{FF4436}
\definecolor{materialorange}{HTML}{FF5722}

\def \billahdebug{}

\ifx \billahdebug \undefined
\newcommand{\fixmesb}[1]{{}}
\newcommand{\fixmeth}[1]{{}}
\else
\newcommand{\fixmesb}[1]{{\bf\textcolor{red}{ [ sB FIXME: #1 ]}}}
\newcommand{\fixmeth}[1]{{\bf\textcolor{blue}{ [ tH FIXME: #1 ]}}}

\fi

\newcolumntype{L}[1]{>{\raggedright\let\newline\\\arraybackslash\hspace{0pt}}m{#1}}
\newcolumntype{C}[1]{>{\centering\let\newline\\\arraybackslash\hspace{0pt}}m{#1}}
\newcolumntype{R}[1]{>{\raggedleft\let\newline\\\arraybackslash\hspace{0pt}}m{#1}}
\begin{document}





\title[A Dataset for Crucial Object Recognition]{A Dataset for Crucial Object Recognition in Blind and Low-Vision Individuals' Navigation}

\author{Md Touhidul Islam}
\affiliation{%
  \institution{Pennsylvania State University}  
  \city{University Park}
  \state{PA}
  \country{USA}
}
\email{touhid@psu.edu}

\author{Imran Kabir}
\affiliation{%
  \institution{Pennsylvania State University}  
  \city{University Park}
  \state{PA}
  \country{USA}
}
\email{ibk5106@psu.edu}

\author{Elena Ariel Pearce}
\affiliation{%
  \institution{Drake University}  
  \city{Des Moines}
  \state{IA}
  \country{USA}
}
\email{elena.pearce@drake.edu}

\author{Md Alimoor Reza}
\affiliation{%
  \institution{Drake University}  
  \city{Des Moines}
  \state{IA}
  \country{USA}
}
\email{md.reza@drake.edu}

\author{Syed Masum Billah}
\affiliation{%
  \institution{Pennsylvania State University}  
  \city{University Park}
  \state{PA}
  \country{United States}
}
\email{sbillah@psu.edu}


\begin{abstract}
This paper introduces a dataset for improving real-time object recognition systems to aid blind and low-vision (BLV) individuals in navigation tasks. The dataset comprises 21 videos of BLV individuals navigating outdoor spaces, and a taxonomy of 90 objects crucial for BLV navigation, refined through a focus group study. We also provide object labeling for the 90 objects across 31 video segments created from the 21 videos. A deeper analysis reveals that most contemporary datasets used in training computer vision models contain only a small subset of the taxonomy in our dataset. Preliminary evaluation of state-of-the-art computer vision models on our dataset highlights shortcomings in accurately detecting key objects relevant to BLV navigation, emphasizing the need for specialized datasets. We make our dataset publicly available, offering valuable resources for developing more inclusive navigation systems for BLV individuals.
\end{abstract}




\maketitle
\section{Introduction}

With the advent of newer algorithms and large-scale, multi-modal datasets, task-specific AI models are converging into a single, task-agnostic, general-purpose model~\cite{brown2020language, raffel2020exploring, Gupta2021GPV, jaegle2021perceiver, jaegle2021perceiver}.
These models are rapidly inching toward their large-scale deployment in various real-world applications.
One specific application from which blind and low-vision (BLV) individuals could benefit is object recognition in real-time.
Through real-time object recognition, blind and low-vision individuals can use these models when they navigate roads and sidewalks.
However, for this to work, the associated model has to be extremely reliable and generate outputs in near real-time.

To this vision, we created a dataset with  21 videos extracted from YouTube and Vimeo featuring BLV individuals navigating outdoor spaces (Section~\ref{sec:taxonomy_dataset}). 
In this dataset, we created a taxonomy of 90 objects with accessibility impacts for BLV individuals, which most mainstream datasets lack. 
We revised this taxonomy by conducting a focus group study with 6 participants who are blind or low-vision or who are orientation and mobility (O\&M) experts (Section~\ref{sec:tax}).
Unlike other taxonomies~\cite{perez2017assessment} for BLV individuals, concepts in our taxonomy are fine-grained and action-oriented. 
For instance, we include \textit{attributes of a sidewalk and driveway} (e.g., accent paving, sloped driveway);
obstructions \textit{likely} to be detected by a white cane (e.g., fire hydrant, gutter); 
obstructions \textit{less likely} to be detected by a white cane (e.g., closed sidewalk, barrier post);
objects that are \textit{too late} to be detected by a white cane (e.g., train tracks);
objects that \textit{pick you before you pick them} (e.g., overhanging tree branches); 
objects that provide \textit{navigational guidance} (e.g., retaining wall, railing, wall, curved railing);
objects \textit{not supposed to be} on the sidewalk (e.g., hose, maintenance vehicle);
moving objects \textit{sharing the sidewalk} (person, bicycle, wheelchair, white cane, guide dog);
\textit{intersections};
objects on the \textit{road shoulder}; objects on the road (e.g., cars, buses, trucks);
traffic signals and street signs (e.g., stop signs);
objects related to building \textit{exits} and \textit{entrances}; and \textit{indoor} objects (e.g., counter, elevator, escalator, stairs, uneven stairs, table, chair).

The suitability of a computer vision model for blind navigation highly depends on its ability to correctly detect objects that are crucial for navigation. 
However, an object recognizer model can only recognize the objects that are present in its training dataset. 
In general, most Object recognition or detection models are trained on datasets such as ImageNet~\cite{deng2009imagenet}, MS COCO~\cite{lin2014microsoft}, Mapillary Vistas~\cite{neuhold2017mapillary}, Kitti~\cite{geiger2012we}, Cityscapes~\cite{cordts2016cityscapes}, Pascal VOC~\cite{everingham2010pascal}, PFB~\cite{pfb_ICCV2017}, and ADE20K~\cite{zhou2017scene, zhou2019semantic}.
Mapillary Vistas is the most advanced open-world dataset that contains high-resolution street-level images and fine-grained annotation of 66 outdoor object categories~\cite{neuhold2017mapillary}. Nevertheless, the ground-truth annotations of Mapillary Vistas and all the other datasets contain significantly fewer objects than the taxonomy in our dataset.

We conduct preliminary evaluations of seven state-of-the-art computer vision models from various genres on our dataset (Section~\ref{sec:assessing_ai_models}).
The results show that these models often fail to generate accurate and consistent predictions within video keyframes.
Most of the shortcomings these models display are because they are unaware of important accessibility-related objects or concepts from our taxonomy, highlighting the concerns about such models being practical for BLV individuals.
We make our dataset publicly available and finish the paper by providing recommendations on how to use our dataset to retrain or build more robust models for blind and low-vision people's navigation.

We summarize our contributions as follows:
\begin{itemize}
    \item A dataset containing 21 navigational videos featuring blind or low-vision individuals. 
    
    \item A taxonomy of 90 objects related to the accessibility of public space, which is refined through a focus group study with 6 participants. 
    
    \item Object labeling for the 90 objects across all 31 video segments from the 21 videos.
\end{itemize}

\section{Background and Related Work}

\subsection{Lack of Fine-Grained Accessibility Annotations in AI Datasets}
Large datasets contain thousands of images annotated with object/class names, bounding rectangles around objects in an image, textual descriptions of an image, questions and answers pertaining to an image, and per-pixel semantic labels.
As the underlying AI tasks evolve---from low-level tasks, such as object classification and detection, to high-level tasks, such as scene understanding, VQA, and semantic segmentation---the effort in annotating data increases, resulting in smaller datasets.
ImageNet~\cite{deng2009imagenet}, for instance, is a classic image classification dataset that contains over 14M images and 1000 objects or classes.  
In contrast, MS-COCO~\cite{lin2014microsoft}, a dataset popular for scene understanding, contains 328K images and 91 object classes.
VQA datasets are a notable exception, partly because it is less demanding to ask free-form questions and answers. For example, VQA v2.0~\cite{balanced_vqa_v2} contains fewer image instances (200K) but contains 614K free-form natural language questions, and 10 answers per question. 
Datasets for semantic segmentation contain several orders of magnitudes smaller since annotating each pixel is laborious and time-consuming. 
For example,  Mapillary Vistas~\cite{mapiliary_iccv17}), where each image contains pixel-wise semantic labels annotated by sighted humans, contains only 25K images and 66 object categories. Other semantic segment datasets, such as KITTI-15~\cite{kitti15_ijcv18} and CityScapes~\cite{cityscapes_cvpr16}, contain fewer images and categories. 

With a few exceptions (e.g.,~\cite{gurari2018vizwiz, saha2019project, theodorou2021disability}), most large-scale, publicly available datasets are not accessibility-aware, i.e., they do not contain annotations important for people with disabilities.
For example, snow over sidewalks, areas near curb cuts, and fire hydrants are important safety information for individuals with blindness who walk with a white cane or with a guide dog. Unfortunately, this perspective does not exist in most mainstream datasets. 
Theodorou et al. reported that computer vision models trained on generic datasets fail to detect objects in  pictures taken by blind users because their pictures tend to be blurrier and might show out-of-frame objects compared with sighted users~\cite{theodorou2021disability}.
KITTI-15~\cite{kitti15_ijcv18} dataset, commonly used to evaluate AI for autonomous driving, lacks many accessibility-aware annotations, such as curb cut, ramp, and crosswalk. 

The lack of accessibility awareness in large-scale public datasets can be consequential, although it is not a surprise. 
Most annotators are sighted crowd workers, who often do not have a nuanced understanding of disability~\cite{tigwell2021nuanced}, who can either ignore individuals with disabilities as ``outliers'' or assign them a generic label (e.g., ``a person with a disability'')~\cite{linton1998claiming, davis2016theDS, morris2020ai}.
However, incorporating accessibility awareness in datasets is challenging. Prior attempts~\cite{theodorou2021disability, park2021designing} at creating disability-first datasets revealed several challenges.
One must engage with target communities, support the communities with accessible data collection tools and annotation frameworks, ensure the quality of the collected data, and strike a balance between the structure and fidelity of data and the demands put on data collectors. 
Because of these challenges, datasets focused on accessibility, such as VizWiz~\cite{gurari2018vizwiz}, contain orders of magnitude fewer annotations than the other similar datasets (e.g., 30K images in VizWiz compared to over 200K images in a common VQA dataset~\cite{balanced_vqa_v2}). 

A possible workaround is to extract data from publicly available video-sharing platforms (YouTube and Vimeo). 
Prior work has studied people with vision impairments by analyzing their YouTube videos~\cite{xie2022youtube, seo2017exploring, seo2021understanding, seo2018understanding}.
We took this approach and extracted 21 videos from YouTube and Vimeo, where BLV users were featured.

\subsection{Convergence of AI}

Convolutional neural network (CNN) architectures such as Mask R-CNN~\cite{he2017mask}, Faster R-CNN~\cite{ren2015faster}, and YOLOv7~\cite{wang2023yolov7} have significantly advanced the field of object detection. These models perform best when the test data closely aligns with their training distribution. However, their effectiveness is constrained by their reliance on predefined object categories, which poses challenges for adaptation to novel object classes or domains.
With newer algorithms and larger, multi-modal datasets, task-specific AI models are converging into a single, task-agnostic, general-purpose model. For example, GPT-4\footnote{https://openai.com/research/gpt-4} is a large multimodal model that can process images, text, and source code.
Foundation models like GPT-4V~\cite{openai2023gpt4,openai2023gpt4vsc,openai2023gpt4vtwa} and LLaVa~\cite{liu2023visual}, trained on diverse internet-scale data, demonstrate remarkable zero-shot generalization in multi-modal visual and language tasks. BLIP~\cite{li2022blip,li2022lavis} enhances multimodal robustness through vision-language pre-training. Advancements in vision foundation models, exemplified by RAM~\cite{zhang2023recognize}, have improved zero-shot image tagging. GPV-1~\cite{gupta2022gpv} is another task-agnostic vision-language model that learns and performs classification, detection, visual question answering, and captioning tasks. 
In general, these models use transformer-based architectures to learn a joint embedding space for different information modalities, such as text and images~\cite{radford2021learning, stroud2020learning, wang2019camp}, text and video~\cite{alayrac2020self, croitoru2021teachtext, miech2020end, wang2021t2vlad}, and audio and video~\cite{alayrac2020self, morgado2021audio}.  
This paper investigates the robustness of general-purpose models, envisioning how BLV users can use them in reality.

Currently, BLV users must use multiple apps as their context changes. Suppose a blind user is walking on the street. To detect an obstacle on the sidewalk, they can use Seeing AI~\cite{SeeingAI2020}, a camera-based app that turns on the rear camera and detects a list of predefined objects. As the person is approaching an intersection, they need to know if the traffic light is green or if the walking sign is on. For this purpose, they may need another app, Oko~\cite{oko_app}, another camera-based app that uses augmented reality and CV to read traffic lights, and walking signage.

They identified that such navigational aids must have two components to facilitate independent mobility: (i) \emph{obstacle avoidance}, and (ii) \emph{wayfinding}~\cite{rafian2017remote}. 
Obstacle avoidance ensures that visually impaired users can move through space safely without running into objects.
Guide dogs and white canes are often used for this purpose. 
Wayfinding, on the other hand, allows them to plan and execute a route to a desired destination.
For wayfinding, having a representation of users' surroundings (i.e., digital maps, cognitive maps~\cite{cognitive_colalge_1993}, building layouts) is essential, and so is \emph{localization}, (i.e., continuously updating their location within that representation).

\subsection{Contextual Relevance in Visual Captioning}

Blind and low-vision users are unable to see the contents of an image. 
Therefore, an alternative option is to create a text-based description of the image. 
This text can later be converted to audio by text-to-speech (TTS) engines or read aloud by screen readers.
The first step in the process is analyzing visual data and translating it into words to create image descriptions~\cite{harper2008web, lazar2004improving, morash2015guiding}. 
Several efforts have worked to improve the convenience of generating image descriptions. 
Many of which concentrate on assisting web application developers~\cite{wai_w3c}. 
Examples include the fundamental guidelines for creating alternative text provided by the Web Content Accessibility Guidelines (WCAG).
Another example would be \emph{The Diagram Center}~\cite{diagram_center_art}, which provides guidelines on understanding the purpose of an image (e.g., decorative, functional), gleaning relevant information from surrounding text, if possible, and judging the age-appropriateness of the generated descriptions.
In addition, guidelines also list the effective attributes of an image caption, including descriptions of foreground, background, color, and the 3D orientation of objects. 
These attributes of a well-formed image caption are in line with the findings from relevant literature~\cite{slatin2003maximum}.

While the aforementioned works focus on standard guidelines for creating image descriptions, other studies suggest that image descriptions should be contextualized based on where the image appears.
In other words, one-size-fits-all solutions would struggle when the context of the image is not taken into consideration. 
Petrie et al.~\cite{petrie2005describing} advocated for this approach but did not present their findings based on individual source types. 
Rather, they proposed guidelines that reflect the common preferences observed across ten different sectors, which recommend descriptions that cover the image's purpose, objects and people, ongoing activities, location, colors, and emotions~\cite{petrie2005describing}.

More recently, recommendations have come out concerning the necessary components for describing images found on social media websites. 
These recommendations include describing all essential objects~\cite{morris2016most}, identifying the individuals in the picture, where it was taken, and how others responded to it~\cite{voykinska2016blind}, indicating significant visual elements, people, and photo quality~\cite{zhao2017effect}, and providing details like facial expression, number of people, age, indoor/outdoor, and nature for people, objects, and settings~\cite{wu2016ask}. In this work, we propose a dataset containing annotations of objects that are essential in the context of blind navigation. This dataset can be beneficial for identifying relevant objects while generating image descriptions or captions.

\section{The Dataset}
\label{sec:taxonomy_dataset}
This section discusses the content of our proposed dataset and the collection procedure.

\subsection{Content of the Dataset}
Our dataset comprises 21 navigational videos from public platforms, each depicting a blind or low-vision individual navigating the road. 
Analyzing these videos, we present a list of 90 objects vital for road navigation for individuals with visual impairments. 
This list was carefully refined through a focus group study. 
Additionally, we provide object labeling for these 90 objects within 31 video segments extracted from the aforementioned 21 videos.
We have made the dataset publicly available \footnote{\href{https://github.com/Shohan29531/BLV-Road-Nav-Accessibility}{https://github.com/Shohan29531/BLV-Road-Nav-Accessibility}}.

\subsection{Video Collection}
\label{sec:video_collection}
We collected free, publicly available videos from two video streaming platforms: YouTube and Vimeo. 
The videos were collected using a systematic keyword-based search. 
Some keywords used in the search were ``blind'', ``vision impairment'', and ``visually impaired'', along with the terms  ``white cane'', ``navigating'', ``orientation and mobility'', and ``training''. 
We kept the duration of the videos between two and twenty minutes long. 
There were no geographical or language limitations in place.
Two researchers watched videos on the top page of the search window and decided whether they were relevant to blind navigation. 
In total, 21 suitable videos were collected, 16 from YouTube and 5 from Vimeo (Table~\ref{table:dataset}). 
The researchers then split the over ten-minute videos into two for easier segmentation and initial analysis. 
For instance, the split videos are V3 \& V4 and V9 \& V10 in Table~\ref{table:dataset}. 
%

\subsection{Identification of Objects with Accessibility Impact}
Two researchers watched all videos and noted when relevant objects first appeared in the video in a Google Sheet. 
An object is deemed relevant 
(a) if it is on the way of the blind individual featured in the video and affects their course (e.g., they changed direction, they collided with the object); 
(b) if it provides meaningful feedback (e.g., produces different sounds, provides different tactile feedback, provides force feedback with the cane); 
(c) if it affects them physically (e.g., they tripped); 
(d) if it could be a hindrance in the future (e.g., the electrical box on a powered gate that sticks out into the sidewalk and could easily be an obstacle to the person using the curb to help orient themselves);
(e) if it could be mislabeled as something else (e.g., the escalator could be mislabeled as stairs which could lead to someone being hurt from not expecting them to be moving); and
(f) if it could be thought of or labeled in the wrong context (e.g., a parked car on the sidewalk could lead a person to believe they were in the street).
%

This process generated around 80 semantic categories related to the accessibility of sidewalks and non-visual navigation. 
We revised these categories by conducting a focus group study with BLV individuals and orientation \& mobility (O\&M) trainers (Section~\ref{sec:tax}).


\begin{table*}[!t]
    \begin{center}
    \small{
        \begin{tabular}{l C{6cm}  C{1.0cm} C{0.50cm} C{0.850cm} C{4.50cm}}
            \toprule
             \textbf{ID} & 
             \textbf{Title}  &
             \rotatebox{90}{\textbf{Duration}} &
             \rotatebox{90}{\textbf{\# Segments}} &
             \rotatebox{90}{\textbf{\# Annotated Seg.}} &
             \textbf{URL} \\
             \toprule
            %
             V1 & Blind Man Walking & 2:24  & 5 & 2 & \url{https://youtu.be/RmsoHyMRtbg}  \\ \hline
             \rowcolor{gray!10} 
             V2 & following a blind person for a day | JAYKEEOUT & 7:02 & 1& 1 & \url{https://youtu.be/dPisedvLKQQ} \\ \hline
             V3 & Orientation \& Mobility for the Blind-1* & 0:00-10:00  & 8 & 2 & \url{https://youtu.be/Gkf5tEbP-oo}\\ 
             \hline
             \rowcolor{gray!10} 
             V4 & Orientation \& Mobility for the Blind-2* & 10:01-19:10 & 4 & 3 & \url{https://youtu.be/Gkf5tEbP-oo?t=602} \\ 
             \hline
             V5 & My First Blind Cane Adventure to Get Coffee | Did I Succeed or Give Up* & 10:00 & 3 & 1 & \url{https://youtu.be/SZM-Le6MEE0} \\ 
             \hline              
             \rowcolor{gray!10} 
             V6 & Using A White Cane | Legally Blind* & 10:00 & 2 & 1 & \url{https://youtu.be/TxUxbXyh7Y4} \\ 
             \hline
             V7 & How a Blind Person Uses a Cane & 4:18 & 4 & 1 & \url{https://youtu.be/xi0JMS1rulo} \\ 
             \hline             
             \rowcolor{gray!10}              
             V8 & Orientation mobility & 9:36 & 2 & 1 & \url{https://youtu.be/6u53Q7IvVIY} \\ 
             \hline                           
             V9 & TAKING THE METRO AND WALKING THROUGH MADRID ALONE AND BLIND-1* & 9:19 & 4 & 1 & \url{https://youtu.be/Vx3-ltp9p-Y} \\ 
             \hline                                        
             \rowcolor{gray!10}  
             V10 & TAKING THE METRO AND WALKING THROUGH MADRID ALONE AND BLIND-2* & 10:00 - 19:00 & 1 & 1 & \url{https://youtu.be/Vx3-ltp9p-Y?t=600} \\ 
             \hline                           
             V11 & Mobility and Orientation Training for Young People with Vision Impairment & 5:48 & 3  & 1 & \url{https://youtu.be/u-3GlbJ5RMc} \\ 
             \hline                           
             \rowcolor{gray!10}  
             V12 & Mobility and Orientation & 8:49 & 4 & 1 & \url{https://vimeo.com/296488214} \\ 
             \hline                           
             V13 & Traveling with the white cane & 2:14 & 3 & 1 & \url{https://vimeo.com/2851243} \\ 
             \hline              
             \rowcolor{gray!10} 
             V14 & Blindness Awareness Month - Orientation and Mobility with ELC and 1st Grade Students & 5:52 & 5 & 2 & \url{https://vimeo.com/758153786} \\ 
             \hline              
             V15 & The White Cane documentary & 5:40 & 3 & 1 & \url{https://vimeo.com/497359578} \\ 
             \hline              
             \rowcolor{gray!10} 
             V16 & Craig Eckhardt takes the subway on Vimeo	& 4:43 & 4 & 1 & \url{https://vimeo.com/17293270} \\ 
             \hline              
             V17 & Guide Techniques for people who are blind or visually impaired* & 10:00 & 3  & 2 & \url{https://youtu.be/iJfxkBOekvs} \\ 
             \hline              
             \rowcolor{gray!10} 
             V18 & Russia: Blind Commuter Faces Obstacles Every Day & 3:20 & 6 & 2 & \url{https://youtu.be/20W2ckx-BcE} \\ 
             \hline              
             V19 & The ``Challenges'' you may not know about ``Blind'' People | A Day in Bright Darkness & 8:00 & 6  & 2 & \url{https://youtu.be/xdyj1Is5IFs} \\ 
             \hline              
             \rowcolor{gray!10} 
             V20 & Blind Challenges in a Sighted World & 3:54 & 5 & 2 & \url{https://youtu.be/3pRWq8ritc8} \\ 
             \hline              
             V21 & What to expect from Orientation \& Mobility Training (O\&M) at VisionCorps & 2:21 & 7 & 2 & \url{https://youtu.be/wU7b8rwr2dM} \\ 
            \bottomrule
    \end{tabular}
    }
    \caption{Summary of the videos in our dataset. We cropped the YouTube videos using \url{https://streamable.com}, which has a crop limit of 10 minutes (max).} 
    \label{table:dataset}
    \end{center}
\end{table*}

\section{Revising Objects with Accessibility Impact: A Focus Group Study}
\label{sec:tax}

We conducted a focus group study with 6 participants to revise our initial list of 80 items and determine whether any objects were missing or redundant.
This section presents the study's methodology, results, and implications for design.

\subsection{Procedure}
\paragraph{Participants.}

Our focus group study had six participants: 
two were blind, two were low-vision, and two were sighted individuals.
The two sighted participants were experts in non-visual navigation because of their professions and proximity to blind individuals. 

Two researchers conducted the study; 
while one researcher asked predefined questions on possibly important objects (from the initial list) for accessible navigation, 
the other researcher took notes and identified new objects of interest.
The session lasted around an hour.
Table~\ref{table:participants} show the participants' demographics, including their age groups, professions, and their familiarity with AI-based applications, such as ``Seeing AI''~\cite{seeingAI}, and ``Oko''~\cite{oko_app}. 
They also use ``Aira''~\cite{Aira} and ``Be My Eyes''~\cite{bemyeyes} services, where remote-sighted humans assist them in video-chat-like communication with smartphones.

\paragraph{Prompts}
We established common ground by describing an imaginary AI app similar to ``Seeing AI''~\cite{seeingAI} and ``Aira''~\cite{Aira} that can detect and read out surrounding objects. 
We generated the following questions/prompts:

\begin{enumerate}
    \item Prompt 1: \emph{Let us imagine we have an AI app that can read out a scene in front of you. What objects would you want the AI to describe proactively?}

    \item Prompt 2: \emph{We have compiled a list of around 80 objects that we felt would be important in navigation for blind individuals. Would you please mention whether an object is relevant or not when we read it out?}

    \item Prompt 3: \emph{Please add more objects that are not on our list but you feel would be important for the scenario (i.e., non-visual navigation). Please also provide a rationale for your choice.} 
    
\end{enumerate}

Apart from the above prompts, we also encouraged the participants to express their opinions on any of the objects if they deemed it necessary.



\begin{table}[t!]
\begin{center}
\small{
\begin{tabular}{l C{1.0cm}  C{1cm} C{6.0cm} C{4cm}}
\toprule

 \textbf{\textsc{ID}} & 
 \textbf{\textsc{Age Group}}  &
 \textbf{\textsc{Gender}} &
 \textbf{\textsc{Profession /Relationship}} &
 \textbf{\textsc{Visual Condition}} \\
 \toprule
 P1 & 45-50 & F  & University Disability Coordinator & Blind (congenital) \\ \hline
 \rowcolor{gray!10} 
 P2 & 40-45 & M  & Orientation and Mobility (O \& M) Trainer & Sighted \\ \hline
 P3 & 50-55 & F  & Non-profit & Legally blind \\ \hline
 \rowcolor{gray!10} 
 P4 & 35-40 & F   & Non-profit & Low-vision  \\ \hline
 P5 & 55-60 & M  & Commercial Truck Driver and the spouse of a blind individual & Sighted  \\ \hline 
 \rowcolor{gray!10} 
 P6 & 55-60 & F  & The Sight-loss Support Coordinator & Low-vision \\
 \bottomrule
\end{tabular}
}
\caption{Focus group study participants' demographics.} 
\label{table:participants}
\end{center}
\end{table}

\subsection{Findings}
\label{subsec:findings}
This section summarizes object-specific opinions from study participants and the relative priority of the objects in our preliminary list. 

\subsubsection{Sidewalk Objects}

Opinions on objects that share the sidewalk (e.g, bicycles, wheelchairs, guard dogs, pets, water hoses) were generally undivided; all participants wanted to be alerted about such objects.
When it came down to relative priority, the participants mentioned bicycles and pets as more important to detect.
According to P1:
\begin{quotation}
\emph{``Bicycles are very important. They are quiet and sometimes you don't know where they are.''}
\end{quotation}

As for the rest of the objects, P5 mentioned that a loose water hose on the sidewalk is a tripping hazard and should be detected beforehand. 
However, P1 and P3 mentioned that their cane would be able to detect a hose and hence would prefer the AI to skip such objects if possible.

\subsubsection{Sidewalk Obstructions}
All sidewalk obstructions (e.g., trees, barrier posts, barrier stumps, pits, trash) were marked as important by our participants, with growing tree branches and barrier posts receiving the highest importance.

\subsubsection{Paratransit and Maintenance Vehicles}
Our participants expressed concerns about maintenance vehicles occupying sidewalks, highlighting the need to detect and avoid such obstacles. 
Responses to paratransit vehicles were mixed, with some participants (P1, P3, and P4) expressing the need for their detection while another participant (P5) found them confusing. 
P5 explained that paratransit vehicles may not be authorized for blind individuals to ride without a pass and stressed the importance of also monitoring their schedule in addition to detecting their presence.

\subsubsection{Traffic Signals and Signs}
There was a general agreement among all of our participants that traffic signals and signs are essential to detect.
However, they agreed that speed limit signs are not necessary for blind individuals.

The participants expressed the need to detect pedestrian crossing signals accurately in case there is an aural warning such as ``audible'' in place.
They mentioned apps such as ``Oko''~\cite{oko_app} that they have explored and used at pedestrian crossings.
Nevertheless, they had major concerns regarding over-reliance on AI tools. P3 said:
\begin{quotation}
\emph{``...Their [Oko app] terms of the agreement is extremely lengthy and nobody ever goes through that. [Looking at us] Make sure you also put all the legal stuff and disclaimers in your future app [visibly concerned about the reliability of such tools]...''}
\end{quotation}

The perception of blind and low-vision individuals regarding the lack of reliability of existing AI tools is a crucial finding.

\subsubsection{Indoor Objects}
Our study participants expressed a strong need for accurate indoor object detection. Specifically, they highlighted the importance of detecting chairs and tables, which can be challenging when searching for empty seating in venues such as restaurants, movie theaters, and public transport such as buses and metro rails. 
Elevator detection was also emphasized due to the frequent difficulty in locating them, with moving walks being another key object to detect, as late detection can lead to injury. 
While escalators and stairs were considered important, they were deemed less critical as they tend to have a significant presence and are typically easier to locate.

\subsection{Revised Taxonomy and Design Implications}
\label{subsec:design_implications}

In this section, we present the revised taxonomy of objects based on the feedback and observations gathered from the study. 
We also discuss the potential design implications for an AI application that caters to the navigation needs of blind and low-vision individuals.

\subsubsection{Newly Identified Objects}
Throughout our study, the participants kept coming up with names of items that they believed to be important in navigation for the blind and low-vision.
Most of the items they mentioned were already on our list, often with slightly different names.
Many new objects came up from the discussion and were subsequently added to our revised taxonomy.
Examples of such objects include doorways, black ice, cobblestone pavements, and moving walks.
The participants found some of the items to be redundant and we subsequently removed those.
We ended up with a final list of 90 objects as shown under different categories in Table~\ref{table:taxonomy}.

\begin{table*}[!ht]
    \centering
\begin{tabular}{c | L{5cm} | C{8cm} } 

  \hline
   \textbf{\textsc{Group}} & 
   \textbf{\textsc{Parent Concept}} & 
   \textbf{\textsc{Accessibility-related objects}}
   \\ 
  \hline
\rowcolor{gray!10} 
1 & Attributes of a sidewalk and driveway & Accent Paving, Driveway (flat), Puddle, Raised Entryway, Sidewalk, Sidewalk Pits, Sloped Driveway, Tactile Paving, Brick Paving, Cobblestone Paving, Unpaved Sidewalk, Wet Surface\\ \hline
2 & Obstructions likely to be detected by a white cane & Fire hydrant, Gutter, Vegetation,
Tree, Brick Wall, Fence, Trash Bins, Lamp Post, Pole, Mailbox\\ \hline
\rowcolor{gray!10} 
3 & Obstructions less likely to be detected by a white cane & Closed Sidewalk, Barrier Post, Barrier Stump, Foldout Sign, Bench\\ \hline
4 & Objects that are too late to be detected by a white cane & Train Tracks, Train Platform\\ \hline
\rowcolor{gray!10} 
5 & Objects that pick you before you pick them & Overhanging Tree Branches\\ \hline
6 & Objects that provide navigational guidance & Retaining Wall, Railing, Wall, Curved Railing\\ \hline
\rowcolor{gray!10} 
7 & Objects not supposed to be on the sidewalk & Hose, Maintenance Vehicle, Trash on Roads, Snow, Water Leakage, Yard Waste, Water Pipes\\ \hline
8 & Moving objects sharing the sidewalk & Person, Bicycle, Wheelchair, Person with a Disability, White Cane, Dog, Guide Dog, Street Vendor \\ \hline
\rowcolor{gray!10} 
9 & Intersection & Pedestrian Crossing, Slopped Curb, Intersection, Crosswalk, Curb, Bridge, Uncontrolled Crossing\\ \hline
10 & Objects on the road shoulder & Road Shoulder, Roadside Parking, Parallel Parking Spot, Paratransit Vehicle\\ \hline
\rowcolor{gray!10} 
11 & Objects on the road & Road, Unpaved Road, Bus, Car, Motorcycle, Road Divider
\\ \hline
12 & Traffic signals and street signs & Traffic Signals, Stop Sign, Sign, Sign Post, Push Button, "Use the Other Door" Sign, Toilet Sign  \\ \hline
\rowcolor{gray!10} 
13 & Objects related to building exits and entrances & Gate, Flush Door, Doorway \\ \hline
14 & Indoor objects & Counter, Elevator, Escalator, Fountain, Stairs, Uneven Stairs, Table, Building, Moving Walk, Pillar, Chair\\ \hline
\rowcolor{gray!10} 
15 & Objects related to public transit & Bus Stop, Turnstile\\ \hline
\end{tabular}
    \caption{Key accessibility-related objects, classified into different groups.}
    \label{table:taxonomy}
\end{table*}

\subsubsection{AI Tools Are Not a Replacement of Physical Assistance Devices}
Our study participants emphasized that AI tools should not replace common physical assistance devices such as white canes.
In the words of P3:

\begin{quotation}
\emph{``I have my concerns about using a phone [all the time] for lack of reliability. Having a physical object such as a white cane to guide is always preferred.''}
\end{quotation}

Our participants reported that while they would appreciate the detection of some objects via AI beforehand, 
they want their cane, which they trust more, to take care of most of the detection tasks.

\subsubsection{Information Priority}
We found a general agreement among our participants that there should be a priority assigned to each object related to navigation. 
For instance, they mentioned that elevators are more important than escalators in an indoor setup, 
as they often have difficulty locating the elevators in large buildings with complex structures.
Among other desired priorities, they mentioned:

\begin{enumerate}
    \item The size of a piece of trash on the sidewalk, if less than four inches, should be ignored. 

    \item The size of any pits on the sidewalk, if small enough to not disrupt the walking experience of a blind user, should be ignored.

    \item Wet surface warnings are redundant, in case it is already raining.

    \item Whether a driveway is flat or sloped is less important. However, the presence of a driveway ahead is a crucial detection task.

    \item Vehicles running on the roads (and not on the sidewalks or shoulders) should not be considered for detection.

    \item If an object is at a distance where a blind user can touch/feel the object with their white canes, they prefer the AI assistance tool to be quiet.

\end{enumerate}

\subsubsection{Configurable Information Presentation}
Our participants desire to be able to choose which objects they wish to be alerted on. 
They also want the AI tool to learn over time from their usage habit, such as turning off a specific object detection that has frequently been opted out by the user in recent times.
As for more advanced customization, they wish the tool to adjust its detection and recommendations based on how they are traveling (e.g., with a guard dog, with a white cane, in a specific weather condition).

P6 wants the tool to tell her how to dress appropriately for the weather, including recommendations for wearing heavy jackets or raincoats.
P4 and P5 believe the tool should generate caution if the temperature and weather in an area are ideal for icing.
All the participants emphasized the importance of detecting ice beforehand, although some of them (P5, P2) acknowledged that it would be very difficult for an AI given how difficult it is for humans with perfect eyesight.
P5 offered an intriguing suggestion concerning ice detection:

\begin{quotation}
\emph{``... Maybe the tool cannot detect icing on its own. But I'm guessing it can have access to the most recent weather data in an area, including warnings such as if the temperature is getting close to the icing point (34-37 degrees Fahrenheit). It should be able to combine the two sources of information to generate a prediction regarding which roads may have icing in the area.''}
\end{quotation}


\subsubsection{Objects that Pick You Before the Cane Picks Them}
Our participants reported certain objects creating noteworthy navigational challenges for blind and low-vision individuals.
For instance, they mentioned that tree branches extending over the sidewalk pose a major threat to them.
Usually existing  at a head-level (or higher) height, tree branches cannot be detected by canes. 
As such, it is quite common for blind individuals to find out about their presence only after they have hit them.
All of our participants agreed that detection of such objects before the objects end up hitting someone is crucial.

Another example of such an object was a train track. In the words of P5:

\begin{quotation}
\emph{``If you detect a train track with your cane and a train is coming, it's already too late... Train tracks are generally tough terrains to navigate; you have all those stones and wooden blocks, making it really easy to get your feet stuck. I would want to be extremely sure that no train is anywhere near when I am stepping into a track.''}
\end{quotation}

Our participants emphasized detecting the aforementioned objects because of the high risk they possess of injuring a blind or low-vision individual.

\subsubsection{Proactive, not reactive.}
An idea that emerged from the focus group discussion was that the individuals could register a set of objects in advance, and the AI app would inform them proactively as the app detects them.
This is consistent with the prior work on remote sighted assistance for blind individuals~\cite{xie2022helping, c4vtochi, lee2022opportunities, lee2020emerging}, which investigated the role of AI in assisting sighted humans so that they can assist blind individuals better. 
AI's roles are to reduce individuals' cognitive load, enhance their ability to stay ahead, and contextualize object detection.
For example, AI should focus more on whether an obstacle is nearby, and if so, where it is located and how it could affect them. 
For example, the impacts of two obstacles---geese waste and an overhanging tree branch--are different and require different measures. 
One requires BLV individuals to avoid hitting the obstacle, and another requires protecting their faces with a hand. 
Thus, having this ability in the app will help BLV individuals plan ahead and reduce surprises.



\section{Ground Truth Labeling and Dataset Analysis}
\label{sec:video_study}

To thoroughly analyze the videos collected through the video collection process mentioned in Section~\ref{sec:video_collection}, 
we split each video into small clips of variable lengths between five and ninety-five seconds.
Each clip revolves around the appearance of objects that are significant in people's navigation on roads and sidewalks.
We refer each clip to a video segment. Table~\ref{table:dataset} shows the number of segments created from each video. 
Using a keyframe extraction tool called \textit{Katna}\footnote{https://katna.readthedocs.io/en/latest/}, we further segmented the clips into keyframes. 
The keyframes are characterized as the representative frames of a video stream that serve a precise and concise summary of the video content, considering transitions in the scene and changes in lighting conditions and activities. 
The number of keyframes (i.e., images) extracted from the video segments was between three and ninety-three. 
Afterward, we manually generated object labeling for object recognition for every keyframe, encompassing all object categories that belong to $L_u$.

\subsection{Object Labeling}
All authors of this paper visually inspected the keyframes to generate ground truth. 
Each author annotated a subset of video segments by observing the changes between two consecutive keyframes. 
For each frame, the existence (denoted by 1) or absence (denoted by 0) of all the objects that belong to $L_{u}$ was annotated. 
If In a keyframe $F_{k}$ the existence of the object $O \in L_{u}$ is denoted by a function $E$, the $E$ can be defined as follows:
\[
    E(O)= 
    \begin{cases}
        1,  & \text{if $O$ exists in $F_k$}\\
        0,  & \text{otherwise}
    \end{cases}
\]
Now the annotation of keyframe $F_k$ can be written as, $(E(O) \ | \ O \in L_u)$. 
Since visually inspecting changes can be subject to ``\textit{change blindness}''---a phenomenon when the visual feed is momentarily interrupted by a blank screen~\cite{simons1997change}---we avoided this by viewing two consecutive keyframe pairs on the screen side-by-side and glancing between them. 

For each video segment, annotating the first keyframe took the longest, typically 5-7 minutes. 
For subsequent frames, we only looked at what objects newly appeared/disappeared from the previous frame and adjusted the annotations accordingly.
Such differential annotation took less than 60 seconds in most cases. 
The annotation task required more time if a new frame appeared with a completely different background or camera viewport. 
With an average of around 15 keyframes per video segment, we needed around $\approx20$ minutes to annotate a video. 
Each video segment was annotated by at least two researchers independently. 
Later, we resolved the conflicts in annotation through collaboration. 

\subsection{Analyzing Annotated Data}
\label{subsec:analyzing_data}
\begin{figure*}[!ht]
    \centering
    \includegraphics[width=1.0\linewidth]{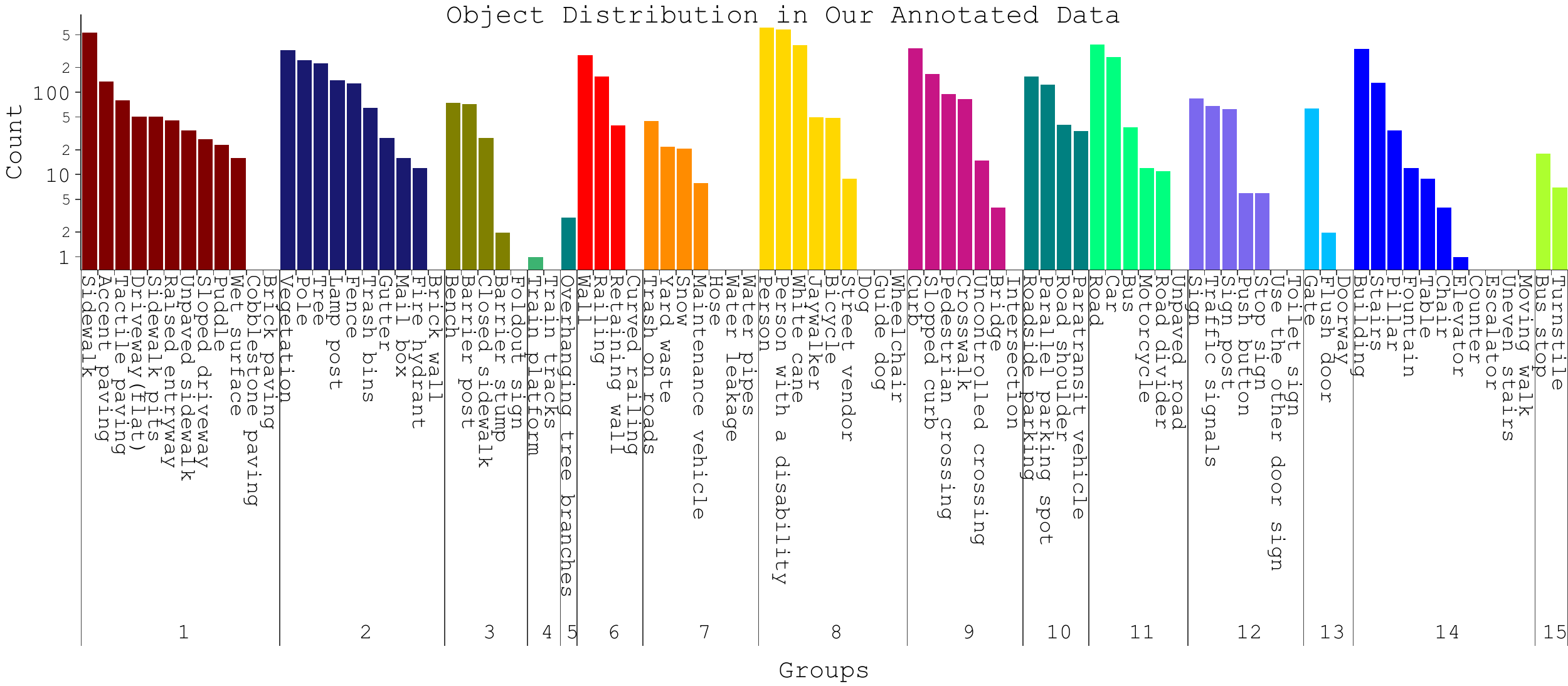}
    \caption{Bar chart representing the object distribution in our annotated data. Each bar represents the number of keyframes in which an object (as labeled on the x-axis) was present. 
    The X-axis also shows the id of the parent concept or group (as described in Table~\ref{table:taxonomy}) to which each object belongs. The Y-axis is in logarithmic scale.}
    \label{fig:dataset_stat}
\end{figure*}

Our annotation encompasses fine-grained object categories (belonging to $L_u$) with which people frequently interact while navigating an outdoor space. 
Some of these objects can create obstacles for people with visual impairment. Therefore, these objects are crucial for outdoor navigation for them. The bar chart in Figure~\ref{fig:dataset_stat} depicts the distribution of objects in our annotated dataset.

According to our study findings, objects of some groups, such as ``\textit{Obstructions less likely to be detected by a white cane (group 3)}'', ``\textit{Objects that pick you before you pick them (group 5)}'', and  ``\textit{Objects not supposed to be on the sidewalk (group 7)}'' are considered as the most significant objects,
since failure in detecting these objects can cause accident or injury to blind people. As shown in Figure~\ref{fig:dataset_stat}, these objects appear in many keyframes of our collected videos. 
Therefore, Accurate detection of these objects is indispensable for blind navigation.

\begin{figure*}[!ht]
    \centering
    \includegraphics[width=1.0\linewidth]{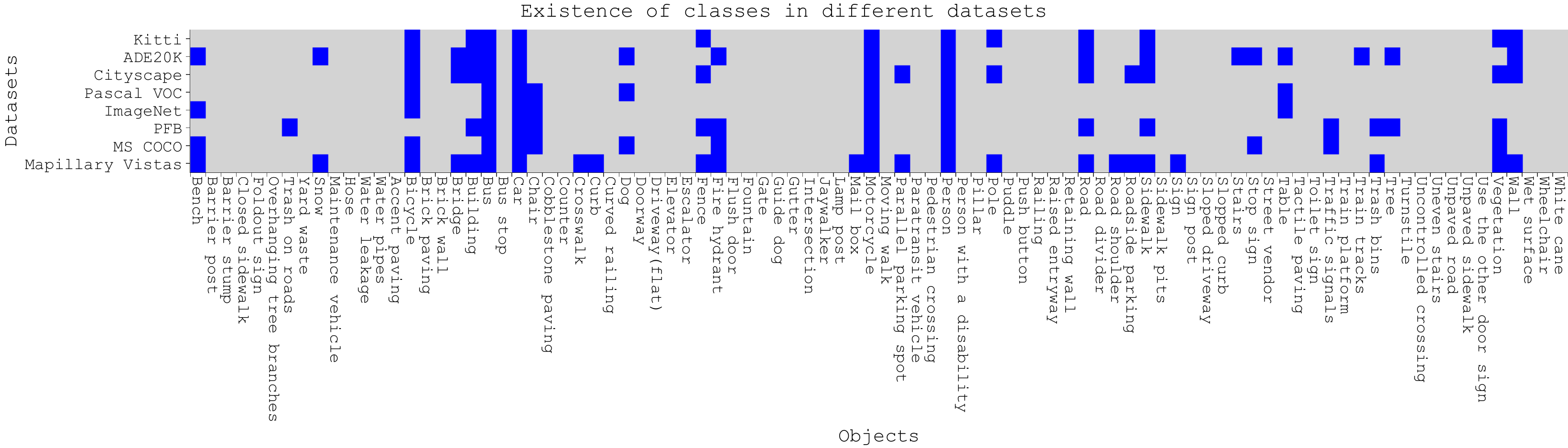}
    \caption{A heatmap representing the existence of different objects of our list $L_u$ in prominent datasets.}
    \label{fig:obj_existance}
\end{figure*}

AI models' readiness for blind navigation highly depends on their ability to correctly detect objects in our proposed list $L_u$ (also shown in Figure~\ref{fig:dataset_stat}). 
However, an object recognizer model can only recognize the objects that are present in its training dataset. 
In general, most Object recognition or detection models are trained on datasets such as ImageNet~\cite{deng2009imagenet}, MS COCO~\cite{lin2014microsoft}, Mapillary Vistas~\cite{neuhold2017mapillary}, Kitti~\cite{geiger2012we}, Cityscapes~\cite{cordts2016cityscapes}, Pascal VOC~\cite{everingham2010pascal}, PFB~\cite{pfb_ICCV2017}, and ADE20K~\cite{zhou2017scene, zhou2019semantic}.
Mapillary Vistas is the most advanced open-world dataset that contains high-resolution street-level images and fine-grained annotation of 66 outdoor object categories~\cite{neuhold2017mapillary}. Nevertheless, the ground-truth annotations of Mapillary Vistas and all the other datasets contain significantly fewer objects of our proposed list $L_u$. 
Additionally, these datasets hardly have annotations for the most significant objects---the ones from group 3, group 5, and group 7; shown within the first nine objects in Figure~\ref{fig:obj_existance} (from \textit{Bench} to \textit{Water Pipes}).
Figure~\ref{fig:obj_existance} shows a heatmap that represents the existence of object categories in some of the most prominent datasets, where 
\colorbox[HTML]{0000FF}{\phantom{x}} in a cell 
means the corresponding object exists in the corresponding dataset. 
In contrast, 
\colorbox[HTML]{D3D3D3}{\phantom{x}} means the object does not exist in the corresponding dataset. 

From Figure~\ref{fig:obj_existance}, we can conclude that most of the objects from our list $L_u$ do not exist in the well-established datasets. 
The first nine objects (from bench to water pipes) from the left of the heatmap are the most significant objects (that belong to groups 3, 5, and 7). 
Only three of these objects exist in some datasets, such as Mapillary Vistas, ADE20K, ImageNet, and PFB.
\section{Preliminary Evaluations on the Dataset}
\label{sec:assessing_ai_models}

We evaluated the preparedness of several state-of-the-art computer vision models for potential application in aiding blind and low-vision individuals utilizing our annotated dataset.
More specifically, we selected models specialized in addressing four distinct computer vision tasks: 
i) object recognition,
ii) object detection,
iii) semantic segmentation,
and iv) visual question answering (VQA). Table~\ref{table:model_type} shows the chosen models alongside their respective computer vision genre. 
We chose a total of seven models from these four types. 
For our experiments, all models were adapted to solve the recognition task.
The adaptation process is explained in Sec.~\ref{subsec:exp_procedure}.

\begin{table*}[!ht]
    \centering
\begin{tabular}{L{6cm} | C{6cm} } 
  \hline
   \textbf{\textsc{Type}} & 
   \textbf{\textsc{Models}}
   \\ 
  \hline
 \rowcolor{gray!10} 
  Recognition Model & RAM (Recognize Anything Model)~\cite{zhang2023recognize} \\
 \hline
  Detection Model & Faster R-CNN~\cite{ren2015faster}, YOLO V7~\cite{wang2023yolov7} \\
 \hline
 \rowcolor{gray!10} 
 Segmentation Model & HRNet V2~\cite{wang2020deep2}, Mask R-CNN~\cite{he2017mask}\\
 \hline
 VQA (Visual Question Answering)~\cite{antol2015vqa} Model & GPV-1~\cite{Gupta2021GPV}, BLIP~\cite{li2022blip, li2022lavis}\\
 \hline
\end{tabular}
    \caption{Selected models for evaluation and their type.}
    \label{table:model_type}
\end{table*}

While a recognition model only predicts the object categories present in a given image, detection models go beyond category prediction and additionally provide rectangular bounding boxes encompassing the objects.
On the other hand, semantic segmentation models detect fine-grained object masks (exact shapes of the objects) along with their categories. 
Traditionally, these models are trained on a set of pre-defined object categories and, hence, can not predict objects that were not present in the vocabulary set of training images.
More recently, the Recognize Anything Model (RAM) can predict open vocabulary categories due to its language decoder~\cite{zhang2023recognize}. 
Finally, VQA models consist of language encoders and decoders, enabling them to predict answers to open-ended questions. 
VQA models can also perform tasks such as image captioning, object localization, and image descriptions. 
Among all the models, RAM and VQA models are suitable for blind and low-vision individuals because of their open-ended capabilities. 

\subsection{Experiments and Observations}
\indent
\subsubsection{Experiment Procedure.}
\label{subsec:exp_procedure}
Recognition models provide a list of predicted objects for a given image, which can be directly used for evaluation. However, other model types provide output in different formats requiring some pre-processing.

\paragraph{Recognition from Detection Model.} Traditionally, detection and segmentation models predict bounding boxes and masks, respectively, for each of the categories. If an object is present in a given image, then the detection model will provide us with the pixel coordinates of bounding boxes and the categories of detected objects. We ignored the bounding boxes in this experiment since we only needed the object categories. 

\paragraph{Recognition from Segmentation Model.} In contrast, the segmentation model predicts a mask for each object category. If an object exists in a frame, its corresponding mask will contain some nonzero values; otherwise, the whole mask will be a matrix of zeros. Thus, We easily created a list of objects that were present in a given image by analyzing the nonzero values of the masks.

\paragraph{Recognition from VQA Model.} For VQA models, we asked them a question for each of the objects regarding whether the object is present or not.
Generally, visual questions are formed from broad categories of question types such as (i) recognition, (ii) common sense, (iii) reasoning, (iv) OCR, and (v) counting. State-of-the-art VQA models perform well when asked visual recognition questions; however, these models are prone to failure when questions from other categories~\cite{li2021adversarial} are asked. To test our selected VQA models, we asked simple visual recognition questions. We generated a set of questions, each focusing on one accessibility-related object as shown in Table~\ref{table:taxonomy}. All of our questions followed a predominant structure like ``Is there object X in the scene?'', where X is an object that came from Table~\ref{table:taxonomy}. Consequently, the answer from the VQA model was either ``yes'' or ``no''. By analyzing the answers, we constructed the list of objects that were in the given image. 
We passed all the extracted keyframes (mentioned in Section~\ref{sec:video_collection}) to all the models we chose for this experiment and generated predictions following the steps mentioned earlier in this paragraph. 
Later, we matched the prediction of each model against our annotation.

\begin{table}[h]
\centering
\begin{tabular}{|c|c|c|c|c|}
\hline
Model & N & Precision & Recall & F1 \\
      &   & (Micro Avg.) & (Micro Avg.) & (Micro Avg.) \\
\hline
BLIP & 90 & 0.3366 & 0.8263 & 0.4783 \\
\rowcolor{gray!10}
GPV-1 & 90 & 0.2273 & 0.8070 & 0.3547 \\
RAM & 54 & 0.8175 & 0.2715 & 0.4077 \\
\rowcolor{gray!10}
YOLO V7 & 12 & 0.9437 & 0.1570 & 0.2692 \\
HRNet V2 & 15 & 0.6110 & 0.3998 & 0.4834 \\
\rowcolor{gray!10}
Mask R-CNN & 12 & 0.6077 & 0.1801 & 0.2779 \\
Faster R-CNN & 12 & 0.6029 & 0.1837 & 0.2815 \\
\hline
\end{tabular}
\caption{Precision, Recall, and F1 score (For all three metrics, higher is better) of all the selected models (shown in Table~\ref{table:model_type}) over our annotated keyframes. $N$ column shows the number of object categories the corresponding model can predict.}
\label{table:metric_summary}
\end{table}

\subsubsection{Results and Observations.}
We calculated the micro average precision, recall, and F1 score of each of the chosen models. 
All the scores are reported in Table~\ref{table:metric_summary}. 
The table also shows the number of objects each model can predict under the $N$ column, out of $L_u$ objects in our dataset. 
From the table, we can observe that the detection and segmentation models can only predict labels for 12 to 15 objects from the list $L_u$---which explains their relatively poor performance.
The lack of representative annotations in their training dataset explains their inability to predict a larger number of object categories.

In contrast, BLIP, GPV-1, and RAM have a built-in language encoder/decoder. 
Therefore, they were able to predict more object categories. 
As we asked questions about the existence of each of the 90 object categories of list $L_u$ to BLIP and GPV-1, they generated answers for each question. 
On the other hand, RAM, being an open-ended object recognition model, could predict 54 object categories from our list. 
BLIP achieved the highest F1 score over 90 object categories among these three models.

\begin{figure*}[!ht]
    \centering
    \includegraphics[width=1.0\linewidth]{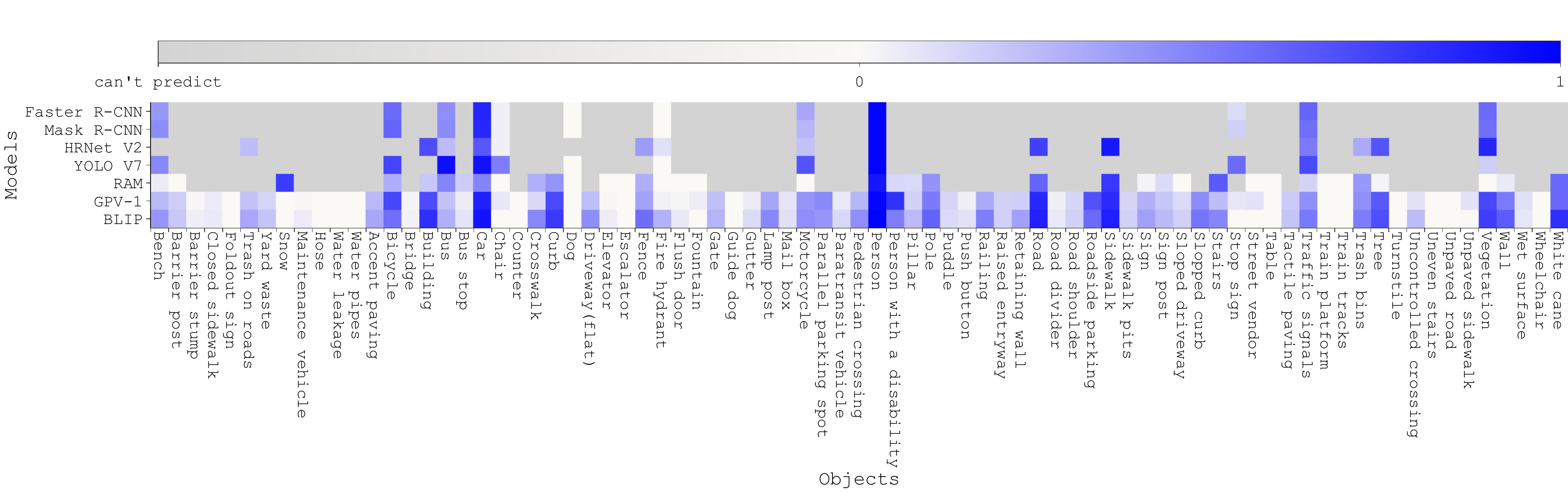}
    \caption{A heatmap representing the classwise F1 score of all the selected models (shown in Table~\ref{table:model_type}).}
    \label{fig:model_comp}
\end{figure*}

We also calculated classwise F1 scores of all the selected models and plotted a heatmap using these scores to visualize the classwise performance of all the models. Figure~\ref{fig:model_comp} shows the heatmap representing the classwise F1 score of all seven models. 
In this figure, \colorbox[HTML]{D3D3D3}{\phantom{x}} in a cell means the corresponding model can not predict the corresponding object.
The F1 score is within the $[0, 1]$ range and is represented by the intensity of the blue color ( \colorbox[HTML]{0000FF}{\phantom{x}} ) in Figure~\ref{fig:model_comp}. 
The heatmap shows that RAM, BLIP, and GPV-1 can recognize more objects than the detection and segmentation models. 
These models can accurately predict a subset of object categories listed in $L_u$, such as persons, cars, buildings, roads, sidewalks, etc. (object columns where the intensity of blue is higher). 

\begin{figure*}[!ht]
    \centering
    
    \includegraphics[width=0.5\linewidth]{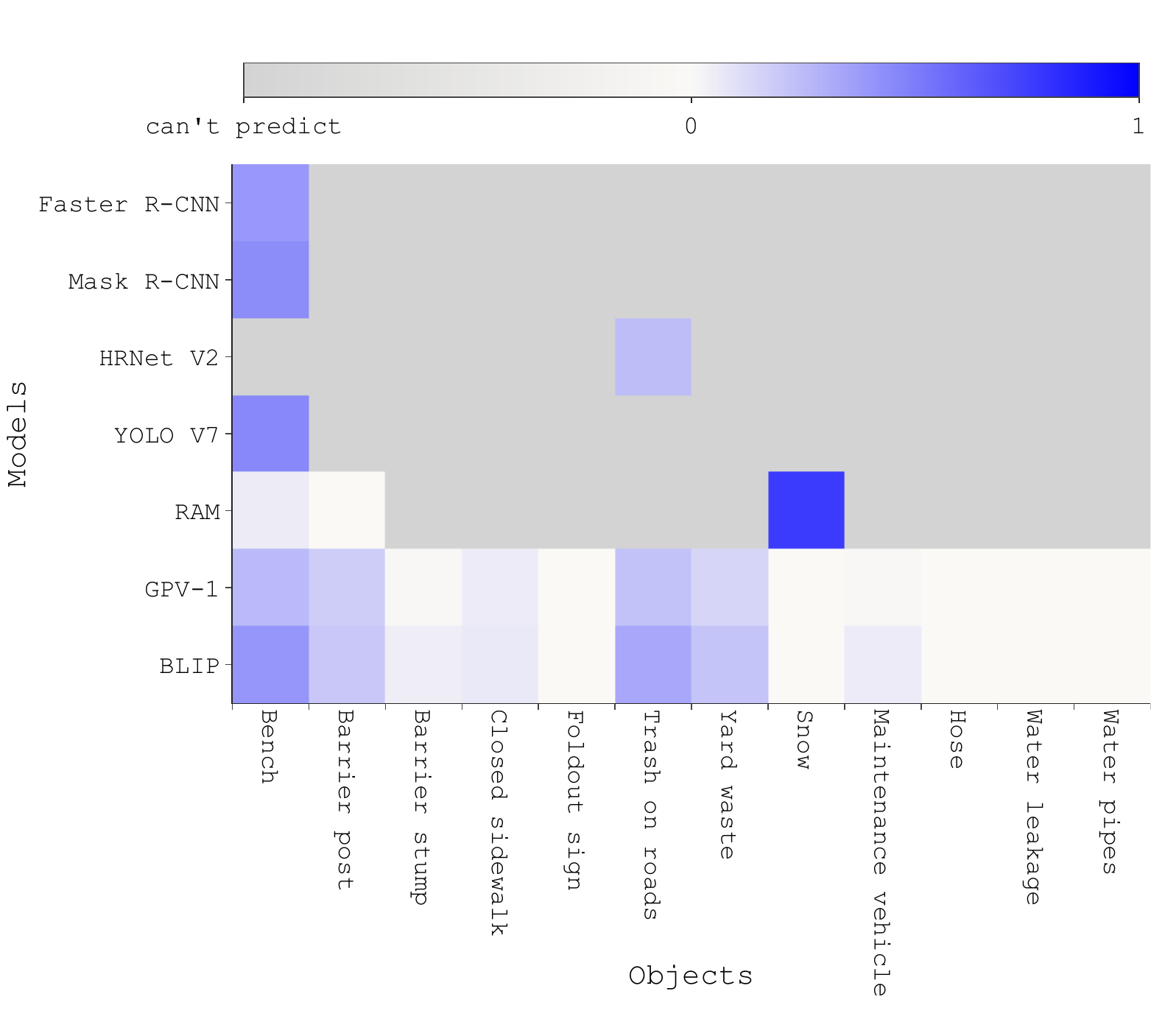}
    
    \caption{A heatmap representing the classwise F1 score of all the selected models for the objects of groups 3, 5, and 7 (shown in Table~\ref{table:taxonomy}).}
    \label{fig:model_comp_dang}
\end{figure*}

Moreover, none of the models can recognize the objects of the most significant groups (groups 3, 5, and 7) well. Figure~\ref{fig:model_comp_dang} represents the F1 scores of all the models for the objects of those three groups. 
The experiments described in this section suggest that none of the selected models are ready to be used in blind or low-vision people's navigation.

\section{Discussion and Implications}
\label{sec:discussion}

This section reflects on how to improve the overall quality of current computer vision models in object recognition for blind and low-vision people's navigation.

\paragraph{\textbf{Incorporating accessibility information in existing datasets.}}

The small base dataset (Table~\ref{table:dataset}) and the comprehensive taxonomy (Table~\ref{table:taxonomy}) we generated can be propagated to any other existing datasets through ``few-shot'' learning~\cite{chen2017deeplab, tan2019efficientnet, bronskill2020tasknorm, finn2017model, vinyals2016matching, zintgraf2019fast}, where AI models can learn to perform VQA tasks on our base dataset. 
Prior work has shown that blind and low-vision individuals can teach a machine (via few-shot learning) to achieve higher accuracy and robustness in real-world applications~\cite{kacorri2017people, lee2019revisiting}.
For example, blind individuals can teach a generic object recognizer to recognize personalized objects, such as their favorite mugs, white canes, or friends' cars, by providing a few, say 5-10, example images or videos of these objects~\cite{kacorri2017teachable}.
As such, our dataset can empower BLV individuals with personalized data-driven navigational technology.

In addition, we can use the knowledge from this paper to understand the accessible impact of individual objects on a scale of 1 to 5.
In the future, we will use this knowledge to develop a deep-learning model that predicts the pixel-wise, contextualized accessibility impact within a scene.
Note that accessible labels are not necessarily the same as an object's name but contain useful and context-rich descriptions.

\paragraph{\textbf{Limitations.}}
Our work has notable limitations. First, we are unaware of the context in which the videos in our dataset were uploaded and whether these were well-rehearsed.
Second, most BLV individuals featuring the videos were experts in O\&M. 
Therefore, their information needs may differ from those not skilled in O\&M. 
Third, we only annotated the presence of 90 objects in each keyframe, which may be inadequate for some AI tasks, such as semantic segmentation.

\section{Conclusion}
While capable, current computer vision models are not yet suitable for direct use in navigation tasks for blind and low-vision individuals. 
This is largely due to their lack of training on objects crucial for navigation in real-world scenarios. 
To tackle this issue, we present a dataset comprising 90 objects vital for road navigation for blind and low-vision individuals. We refined this list through a focus group study involving both blind and low-vision individuals and sighted companions of blind individuals. 
Additionally, we collected 21 navigational videos from publicly available sources and provided object labeling for 31 segments of these videos. 
Our dataset fills a significant gap in existing datasets, which typically lack many of the objects crucial for navigation identified through our research. 
This inadequacy is underscored by the subpar performance of seven state-of-the-art vision language models observed in our preliminary evaluation.
By making our dataset publicly available, we aim to facilitate the retraining and enhancement of existing computer vision models for real-time detection of road objects, ultimately improving navigation for blind and low-vision individuals.

\section{Ethical Considerations}
As per the Common Rule of the federal regulations on human subjects research protections (45 CFR 46.104(d)(4)), the collection of existing data is exempt from IRB review when the sources are publicly available. Nonetheless, we reached out to the video uploaders on YouTube and Vimeo to give them credit and to solicit their permission.

\bibliographystyle{ACM-Reference-Format}
\bibliography{Bibliography, Bibliography2, Bibliography3}

\clearpage

\appendix

\renewcommand{\thefigure}{A\arabic{figure}}
\setcounter{figure}{0}

\renewcommand{\thetable}{A\arabic{table}}
\setcounter{table}{0}

\end{document}